\documentclass{article}
\usepackage{arxiv}
\usepackage[utf8]{inputenc} 
\usepackage[T1]{fontenc}    
\usepackage{hyperref}       
\usepackage{url}            
\usepackage{booktabs}       
\usepackage{amsfonts}       
\usepackage{nicefrac}       
\usepackage{microtype}      
\usepackage{lipsum}		
\usepackage{graphicx}
\usepackage{natbib}
\usepackage{doi}
\usepackage{float}

\title{Deep VULMAN: A Deep Reinforcement Learning-enabled Cyber Vulnerability Management Framework}

\date{}

\author{ \href{https://orcid.org/0000-0002-9326-291X}{\includegraphics[scale=0.06]{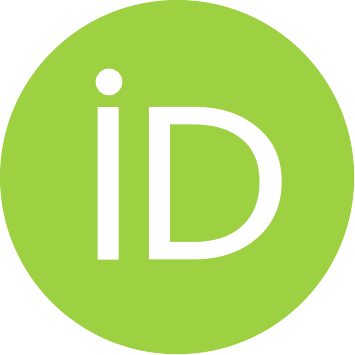}\hspace{1mm}Soumyadeep Hore}\\
	Industrial and Management Systems Engineering\\
	University of South Florida\\
	Tampa, FL 33620 \\
	\texttt{soumyadeep@usf.edu} \\
	\And
	{\hspace{1mm}Ankit Shah} \\
	Industrial and Management Systems Engineering\\
	University of South Florida\\
	Tampa, FL 33620 \\
	\texttt{ankitshah@usf.edu} \\
    \And
	{\hspace{1mm}Nathaniel D. Bastian} \\
	Army Cyber Institute\\
	United States Military Academy, West Point\\
	NY 10996 \\
	\texttt{nathaniel.bastian@westpoint.edu} \\
}

\renewcommand{\shorttitle}{Deep VULMAN}

\hypersetup{
pdftitle={Deep VULMAN: A Deep Reinforcement Learning-enabled Cyber Vulnerability Management Framework}, 
pdfsubject={},
pdfauthor={Soumyadeep Hore, Ankit Shah, Nathaniel D. Bastian},
pdfkeywords={Cyber Vulnerability Management, Vulnerability Prioritization, Security Resources Optimization, Deep Reinforcement Learning, Integer Programming, DRL Cyber Framework.},
}

\begin{document}
\maketitle

\begin{abstract}
Cyber vulnerability management is a critical function of a cybersecurity operations center (CSOC) that helps protect organizations against cyber-attacks on their computer and network systems. Adversaries hold an asymmetric advantage over the CSOC, as the number of deficiencies in these systems is increasing at a significantly higher rate compared to the expansion rate of the security teams to mitigate them in a resource-constrained environment. The current approaches are deterministic and one-time decision-making methods, which do not consider future uncertainties when prioritizing and selecting vulnerabilities for mitigation. These approaches are also constrained by the sub-optimal distribution of resources, providing no flexibility to adjust their response to fluctuations in vulnerability arrivals. We propose a novel framework, Deep VULMAN, consisting of a deep reinforcement learning agent and an integer programming method to fill this gap in the cyber vulnerability management process. Our sequential decision-making framework, first, determines the near-optimal amount of resources to be allocated for mitigation under uncertainty for a given system state and then determines the optimal set of prioritized vulnerability instances for mitigation. Our proposed framework outperforms the current methods in prioritizing the selection of important organization-specific vulnerabilities, on both simulated and real-world vulnerability data, observed over a one-year period.
\end{abstract}


\footnote{This work has been submitted to Elsevier for possible publication. Copyright may be transferred without notice, after which this version may no longer be accessible.}

\section{Introduction}
Adversaries are actively looking to exploit unpatched vulnerabilities in the computer and network systems to cause significant damage to public and private organizations. Recently, the United States White House issued a memo urging organizations to \emph{promptly identify and remediate vulnerabilities} in their systems, among other recommendations to bolster cybersecurity against the adversaries~\cite{WH}. Major challenges faced by the organizations to implement this recommendation result from a significant recent increase in the number of new vulnerabilities that are reported in the National Vulnerability Database~\cite{NVD}, as well as the lack of security personnel (resources) available to mitigate them. This has resulted in vulnerabilities persisting in the computer and network systems of the organizations for a long time, thereby creating a significant advantage for the adversaries. There exists a critical gap in research needed to develop resource-constrained approaches for effectively identifying and mitigating important organization-specific security vulnerabilities to protect against adversarial exploitation and minimize damage from cyber-attacks.

A typical cyber vulnerability management process starts with the scanning of the software and hardware components of an organization’s network with a vulnerability scanner (such as Tenable, Qualys, or IBM) to find vulnerabilities reported in the NVD. The generated vulnerability report contains all vulnerability instances found in the network along with their attributes, which include the common vulnerability exposure (CVE) code, host name, description, and the common vulnerability scoring system (CVSS) severity rating, among others. The security teams at the cybersecurity operations centers (CSOCs) then assign resources to mitigate the vulnerability instances based on certain schemes. Examples of actions taken by security personnel are applying patches (vendor-supplied or CSOC-designed), upgrading software, disabling services, and adding IP filters, among others. The current approaches for vulnerability management, which include methods employed at the CSOCs and proposed in recently published literature, use rule-based mechanisms or static (one-time) optimization models~\cite{farris2018vulcon, shah2019vulnerability, hore2022towards} to prioritize the selection of vulnerabilities for mitigation, given the number of resources available at a particular time-step (for instance, a week or a month). 

There are many shortcomings in the current approaches. First, the vulnerability selection process does not include a comprehensive list of factors associated with the host machine and the respective organizational environment to determine the true priority of a vulnerability instance found in a scan report. For instance, a CSOC security team performs many functions, which include intrusion detection system (IDS) alert management along with vulnerability management. An IDS alert log can identify host machines with possible intrusion attempts and integrating this information, along with other factors such as the CVSS severity score, into prioritizing vulnerability instances found on such machines can help better protect against potential attacks. Second, recently proposed optimization models have focused on selecting vulnerability instances from dense reports to maximize their cumulative vulnerability utility or exposure score, given a limited number of available resources. Such an approach does not result in the selection of all important vulnerabilities as these mathematical formulations focus on the value of selecting a vulnerability instance based on the time it takes to patch or mitigate it. These methods will select a larger number of less important vulnerabilities if their mitigation time is considerably low when compared to an important vulnerability with a significantly higher mitigation time. Third, the current approaches assume a deterministic environment for solving this problem, in which the number and type of vulnerability arrivals are considered to be known and are uniformly distributed across the time horizon. They do not take into account the uncertainty in vulnerability arrivals and consider a pre-determined (often, an equal) number of resources distributed across all the individual decision-making time-steps to prioritize the selection of vulnerabilities for mitigation.

Cyber vulnerability management is a continuous process aimed at strengthening the security posture of an organization within an infinite time horizon. This requires sequential decision-making, and to make it robust against the uncertainties in the process, it is imperative that (i) the number of resources to be allocated at each time-step is optimized and (ii) the important vulnerabilities are identified and prioritized for mitigation, given the optimized allocation of resources. Our research objective is to fill the current gap in the cyber vulnerability management process by proposing a novel artificial intelligence (AI) enabled framework, powered by a deep reinforcement learning (DRL) agent and an integer programming method for effective vulnerability triage and mitigation.

The main contributions of the paper are as follows. First, we developed a novel dynamic cyber vulnerability triage framework, Deep VULMAN, which is designed to combat the uncertainty in the vulnerability management process and select the most important vulnerability instances for mitigation from a dense list of vulnerabilities identified in the network. Unlike other methods in recent literature, we pose the problem as a sequential decision-making problem and segregate the vulnerability management process in our proposed framework into two parts: (i) determining the near-optimal amount of resources required for mitigation, given the observed state of the system and (ii) determining the optimal set of prioritized vulnerability instances for mitigation which has the maximizing average cumulative attribute score among all the vulnerability instances. Second, we developed a DRL agent based on a policy gradient approach that learns to make near-optimal resource allocation decisions under uncertainty in vulnerability arrivals. The agent continuously interacts with a simulated CSOC operations environment built using real-world vulnerability data and gets feedback from a novel reward signal engineered from (i) the mitigation of important vulnerabilities and (ii) the number of resources utilized at each time-step. Third, we formulated and solved a combinatorial mathematical model with an integer programming method for vulnerability prioritization and selection for mitigation with the allocated resource decision from the DRL agent. Unlike the recent methods in the literature, we present a unique formulation that generates an optimal set of prioritized vulnerability instances for mitigation, which has the maximum average cumulative attribute score among all the vulnerability instances. Fourth, to the best of our knowledge, this study is the first to propose a framework that integrates alert information from IDS to vulnerability data to improve the vulnerability management process at a CSOC. This is a major step toward building a robust defense system against adversaries. Our experiment results demonstrated that with this added information from the alert logs, through prioritized vulnerability instances, we were able to find machines that had very old or expired versions of software making them easier targets for the adversaries. Finally, we provided valuable insights obtained using our proposed framework by comparing our results with recent vulnerability prioritization and selection methods from the literature. Our experimental results using real-world vulnerability data show that our approach is more efficient and effective in terms of selecting important organization-specific vulnerabilities in comparison with the other methods.

The paper is organized as follows. Section 2 presents the related literature. Section 3 presents the proposed Deep VULMAN framework, which consists of the CSOC operations simulation environment and the AI-enabled decision-support component that recommends near-optimal decisions for vulnerability management. Section 4 presents the numerical experiments performed using real-world vulnerability scan data. Section 5 presents the experimental results and comparisons with recent methods from the literature. Lastly, in Section 6, we provide conclusions.

\section{Related Literature}
We organized the literature review by dividing the related literature into two topics: (i) vulnerability scoring systems and triage methods, and (ii) DRL approaches in solving sequential decision-making problems under uncertainty. 

\subsection{Vulnerability Scoring Systems and Triage Methods}
To gauge the severity or threat of a vulnerability, it is important to have a mechanism for scoring the attributes or impacts of the vulnerability. In 2006, ~\cite{mell2006common} proposed the common vulnerability scoring system (CVSS) to provide a base score to quantify the vulnerability severity. Later, in 2007, the same authors proposed CVSS version 2 to cover the shortcomings of CVSS version 1 by reducing inconsistencies, providing additional granularity, and increasing the capability to reflect a wide variety of vulnerabilities~\cite{mell2007complete}. The CVSS framework is managed by the Forum of Incident Response and Security Teams (FIRST), and the latest version of CVSS in use today is version 3.1. The CVSS metric consists of eight base metrics, three temporal metrics, and four environmental metrics~\cite{FIRSTcvss3}. However, the computation of environmental metrics is complicated and not well proven~\cite{gallon2010impact}. The NVD omits the temporal and environmental metrics and considers only the base metrics when calculating the CVSS severity of reported vulnerabilities~\cite{fruhwirth2009improving}. CVSS base metric group is a common choice of application among most organizations to gauge the severity of the vulnerabilities present in their network. However, anecdotal and literary evidences suggest that the CVSS base score alone is not sufficient to measure the impact of a vulnerability in a particular organization due to the absence of organizational context~\cite{fruhwirth2009improving,farris2018vulcon,holm2011quantitative,holm2012empirical}. There have been many contributions from researchers to bridge this gap. Some of the important contributions are by \cite{mcqueen2009empirical}, in which the authors proposed two metrics Median Active Vulnerabilities (MAV) and Vulnerability-Free Days (VFD) based on the report time of the vulnerability and time when the patch is issued by the vendor~\cite{allodi2014comparing}; they considered black-market exploit data to boost the statistical significance of the indication pertaining to the true severity of a vulnerability; \cite{farris2018vulcon} proposed two performance metrics: Total Vulnerability Exposure (TVE) that scores the density of unmitigated vulnerabilities per month and Time-to-Vulnerability Remediation (TVR) based on the maximum amount of time (in months) an organization is willing to tolerate the presence of a certain vulnerability in their system; and \cite{hore2022towards} presented a novel Vulnerability Priority Scoring System (VPSS) that takes into account the context of the vulnerability along with the CVSS score by considering relevant host machine information (positional significance of the host machine, level of importance of the host machine, and protection level of the host machine).
\subsection{Deep Reinforcement Learning (DRL) Approaches}
DRL is one of the most promising solution methods for obtaining near-optimal policies under uncertain (stochastic) conditions. DRL was first applied by Mnih et al. in 2013 to successfully learn a control policy from sensory inputs with high dimensions~\cite{mnih2013playing}. Today, DRL has been used in various application domains such as autonomous vehicles, stock trading, robotics, cyber-security, and marketing, among others~\cite{bogyrbayeva2021reinforcement,kirtas2020deepbots,liang2020precision}. The model-free DRL methods in published literature can be broadly classified in two parts: value-based and policy-based. In value-based DRL approaches, we try to estimate the Q-value or a state-action pair by employing a deep neural network estimator. Policy based methods aim to directly learn the stochastic or deterministic policies, where the action is generated by sampling from the policy. Mnih et. al proposed a novel method, Deep Q Learning (DQN), which is a value-based method with superior performance demonstrated on Atari 2600 games. Some of the notable advancements made in the area of value-based DRL methods include the works by: Van hasselt et. al, who proposed DRL with double q-learning (DDQN) to overcome the overestimation suffered by DQN~\cite{van2016deep}; and Wang et. al, who proposed the dueling network architectures for DRL with two identical but separate neural network estimators for estimating the state value function and action advantage function~\cite{wang2016dueling}, among others. One of the popular advancements in policy-based methods includes the work by Mnih et. al, who presented asynchronous methods for DRL with parallel actor learners, asynchronous advantage actor critic (A3C), and outperformed others on Atari 2600 games~\cite{mnih2016asynchronous}. Vanilla policy gradient algorithms generally suffer from high variance, poor sample efficiency, and slow convergence. \cite{schulman2015trust} presented Trust Region Policy Optimization (TRPO) that limits the policy update with a certain KL-divergence constraint and also guarantees monotonic improvement. In 2017, \cite{schulman2017proximal} proposed Proximal policy Optimization (PPO) that has all the advantages of TRPO, and in addition, it is simpler, faster, and more sample efficient. PPO uses a clipped surrogate objective function that prevents large changes in the policy. The clipped surrogate objective is also a lightweight replacement of the KL-divergence constraint in TRPO. Due to its simplicity, sample efficiency, and robustness to hyper-parameter tuning, PPO is a promising approach to solving dynamic sequential decision-making problems.

There is a clear gap in the literature for cyber vulnerability prioritization and selection, as most of the work has been focused on formulating one-time (static) strategies for selecting vulnerabilities from dense vulnerability reports by considering a fixed amount of resource availability and without taking future vulnerability arrivals into account. To the best of our knowledge, no research has addressed the vulnerability management problem as a sequential decision-making problem under the uncertainty of vulnerability arrivals and with resource fluctuations. This paper focuses on strengthening the security posture of the CSOCs by generating robust vulnerability management policies for real-world uncertain environments. Next, we present the proposed framework for dynamic vulnerability management under uncertainty.

\section{Deep Reinforcement Learning-enabled Cyber Vulnerability Management (Deep VULMAN) Framework}

\begin{figure}[H]
\centering
\includegraphics[width=0.8\textwidth]{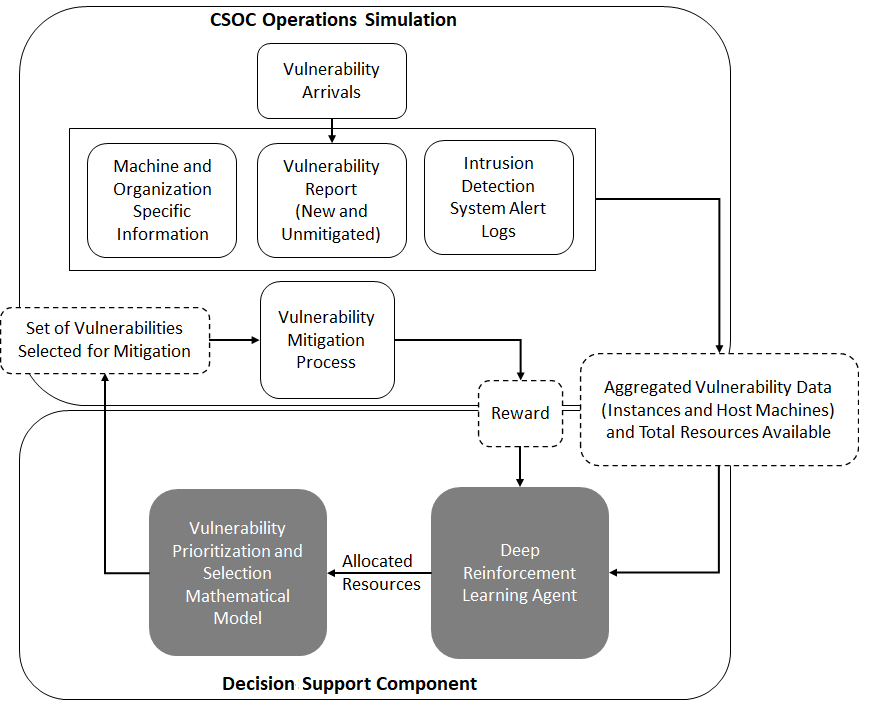}
\caption{Deep VULMAN Framework for Cyber Vulnerability Management.}
\label{frame}
\end{figure}

We propose the development of a sequential decision-making framework that provides a dynamic resource allocation strategy along with an optimal selection of vulnerabilities that are prioritized for mitigation. Figure~\ref{frame} shows a schematic representation of the proposed Deep VULMAN framework. The framework consists of two key components: (i) a CSOC operations environment, where relevant computer and network data are collected and aggregated using various software applications, and (ii) a decision-support component, in which (a) a DRL agent is trained using a policy gradient algorithm to make near-optimal resource allocation decisions under uncertainty and (b) an integer programming model is developed to generate the set of vulnerabilities, which are prioritized for mitigation with the amount of resources allocated by the DRL agent. We first describe the CSOC operations environment, in which we propose a simulator to overcome the data insufficiency issues for training the DRL agent, followed by the decision-support component.

\subsection{CSOC Operations Simulation}
Obtaining a real and large data set for a research study is a major challenge for cybersecurity researchers. Very few studies in published literature, such as \cite{farris2016preliminary}, \cite{xu2018modeling}, and \cite{shah2019vulnerability}, have investigated the process of cyber-incident or vulnerability emergence using historical data. However, these have been small and/or private data sets. The unavailability of data sets is due to a lack of complete information in a cyber environment or confidentiality reasons. Cyber-incident data have been studied in \cite{haldar2017mathematical} and \cite{kuypers2016department} for large cyber breaches and it has been found that a Poisson distribution provides the best fit for describing the arrivals in these data sets. It is imperative that the DRL agent interacts with an environment that closely resembles the real CSOC operations to learn the best policies that can be implemented in real-world conditions. Hence, to overcome the challenges, such as having insufficient data to properly train a DRL agent or learning in a slow-moving real-world environment~\cite{dulac2019challenges}, we built a simulator from the large amount of real data that we collected by working with a CSOC. We developed an agent-based discrete event simulation (DES) algorithm with fixed-increment time progression to model the vulnerability management process at a CSOC. The agent-based approach is added to the traditional DES to accommodate the interaction between the DRL-agent (explained in the next section) and the simulation environment. The inputs to the algorithm are the various vulnerability scan reports and other relevant network related information obtained from applications such as Nessus, Lansweeper, and IDS. The uncertainty in the vulnerability arrival process is captured in the simulator by randomly generating vulnerability arrival patterns at each time-step. For example, there could be a high, medium, or a low number of vulnerability arrivals in a given week. Vulnerability instances with varying characteristics and related host machine data are randomly sampled from the historical data sets at each time-step. The arrivals are generated using a Poisson distribution with varying mean, which can be obtained from the historical data.

The cyber vulnerability instances, the respective host machine information, and the resources available are then passed on to the decision-support component (see Figure~\ref{frame}) as the state of the system at the given time-step. The action pertaining to this system state is then taken as input by the simulated environment from the decision-support component. The simulation algorithm executes this action, which contains the set of vulnerabilities selected for mitigation. A scalar reward is computed in the simulator, which consists of two terms, one related to the mitigation of important vulnerabilities and another for the number of resources utilized. The details of the action selection and the reward function are presented in the next section. The cumulative time taken to mitigate the selected vulnerabilities is then deducted from the total available time at the beginning of the time-step. The environment is then stepped forward to the next time-step with the remaining resources. A new set of vulnerability instances is then generated, and this process continues for the entire episode (e.g., a month). The simulator adds the new set of vulnerabilities (arrivals) to the unmitigated set of vulnerabilities from the previous time-step. 

\subsection{Decision Support for Vulnerability Management}
The objective of this research is to identify and prioritize important cyber vulnerabilities for mitigation under uncertainty of future vulnerability arrivals in a resource-constrained system. It is to be noted that the decision-making problem can be broken down into obtaining two decisions: (i) determining the near-optimal resources to be allocated and (ii) determining the set of vulnerability instances for mitigation given these resources, which reduces the vulnerability exposure of the organization in the long run. The former decision of allocating the appropriate amount of resources is affected by the uncertainty in the environment and the CSOC can enhance their security with a dynamic resource allocation strategy. Once the decision on the amount of resources allocated is made, the mathematical model can be invoked to optimally select the set of important vulnerabilities for mitigation. We first describe the DRL problem formulation for optimizing the resource allocation strategy, followed by the formulation of the mathematical model, which outputs the vulnerability selection decision.

\subsubsection{DRL Formulation}
\label{sec}
The problem of making sequential decisions for resource allocation to mitigate important vulnerabilities and thereby reducing the vulnerability exposure and strengthening the security posture of an organization in the long run can be formulated as a Markov decision process (MDP). The key elements of the MDP formulation are as follows:

\begin{itemize}
    \item \textbf{State}, $s_t$, represents the information that is visible to the agent at time $t$, which consists of the vulnerability instances, their respective attributes, and the total amount of resources available. The state space is $N * (M+1)$ dimensional, where  $M$ is the number of attributes and $N$ is the maximum number of vulnerabilities historically found in the vulnerability scan reports. We use the concept of zero padding to fill empty rows ($N$ - $J$ number of vulnerabilities found at each scan) with zeros~\cite{lin2020softgym}. The state space provides the DRL agent with the information needed to make the resource allocation decision for vulnerability selection.
    \item \textbf{Action}, $a_t$, represents the control. The action is the amount of resources to be allocated at time $t$, given a state, $s_t$. The action space is continuous for this problem.
    \item \textbf{State transition function} determines the probability with which a system will transition from state $s_t$ to $s_{t+1}$ under action $a_t$. The state transition probabilities for this problem are unknown and the possible number of state transitions are very high (state space explosion). Hence, it is infeasible to determine the state transition probabilities.
    \item \textbf{Reward}, $r_t$, is a measure of the goodness of an action, $a_t$, taken in a given state, $s_t$, at time $t$. The agent’s goal is to maximize the long-term cumulative reward. Hence, setting up the reward signal is critical to train the agent to achieve the research objective. In this research, we engineer a novel reward function, which consists of two weighted terms. The reward is obtained from: (i) the mitigation of important vulnerabilities ($r^1$) and (ii) the number of resources utilized ($r^2$). The reward function, at time $t$, is given by Equation~\ref{eq2} where $w_1$ and $w_2$ are weights associated with the reward terms and whose sum must be equal to 1.

\begin{equation}
\label{eq2}
r_t = w_1 * r^1_t + w_2 * r^2_t
\end{equation}

The importance of a vulnerability instance is determined by taking into consideration the following attributes: asset criticality, level of protection, and organizational relevance of the host machine, the CVSS severity of the vulnerability instance, and if the host machine has been identified in any IDS alerts. These attributes are obtained using various applications from the organization’s computer and network systems. Categorical attributes are transformed into numerical values based on certain rules from literature. We use the same scheme, as in \cite{hore2022towards, shah2018two, farris2018vulcon}, to identify various categories for each attribute and assign normalized numerical values. For completeness, we describe this scheme here. The following attributes: asset criticality, level of protection, and organizational relevance of the host machine are assigned either of the three categories, high, medium, or low. It is to be noted that more categories could be added to this list such as a critical priority category. The categorical attribute with the highest priority is assigned a numerical value of 1 and the lowest is assigned a value of 0.1. The ordered categories in between the highest and lowest priorities are then assigned numerical values based on a linear scale. For instance, if the asset criticality associated with a certain machine is of the highest priority (critical), then it takes a value of 1 and if it is the lowest priority (low), then it is assigned a value of 0.1. The CVSS severity score for a vulnerability is obtained from the NVD through the application, which is then normalized between 0 and 1. For instance, NESSUS provides this score as a part of the scan report, which ranges from 1 to 10. If the machine is identified in the IDS alert logs for a possible intrusion, then the attribute is assigned a value of 1, else 0. All these factors are considered equally important. There is a positive reward for selecting vulnerabilities, which is calculated by taking the average of all the attribute values of the selected vulnerabilities. If there are $J$ number of selected vulnerabilities and  $v_{ij}$ represents the value of attribute $i$ of the vulnerability instance $j$, then the positive reward can be calculated as $r^1_t = \frac{\sum_{j=1}^{J}\sum_{i=1}^{I} v_{ij}}{I*J}$. Since the CSOC operations environment is resource constrained, there exists a trade-off between the number of vulnerabilities selected for mitigation and the number of resources that remain available for vulnerability selection in the next time-step. Hence, we assign a small cost to the utilization of the resources in this formulation. For the $J$ number of vulnerability instances selected for mitigation with $S_j$ representing the time required to mitigate vulnerability $j$ and $C$ representing the cost/unit resource utilized, then the resource utilization penalty ($r^2_t$) is calculated as $-\sum_{j=1}^{J} C*S_j$.
\end{itemize}

The large state space and continuous action space make this problem infeasible to solve using conventional reinforcement learning approaches. To overcome the issue of not being able to calculate and store the action-value (or Q value) for all possible state-action pairs due to state space explosion, we propose a deep neural network-based learning model with a policy gradient algorithm for efficiently solving this problem~\cite{silver2014deterministic}. Vanilla policy gradient algorithms have disadvantages such as poor data efficiency, lack of robustness, and are often subjected to large changes in policies resulting in unstable learning. Hence, we propose the proximal policy optimization (PPO) approach~\cite{schulman2017proximal} for solving this problem, which is an on-policy algorithm that overcomes the aforementioned challenges. PPO ensures smoother learning of the policies with the objective clipping feature. Additionally, PPO is easy to implement and tune, and provides better sample efficiency.

\subsubsection{Vulnerability Prioritization and Selection Model}
The prioritization and selection of cyber vulnerability instances is achieved by solving a mathematical model, whose solution provides us with the set of prioritized vulnerabilities selected for mitigation by the available resources (decision made by the DRL agent). The vulnerability selection problem is posed as a combinatorial optimization problem and solved using integer programming. The static vulnerability prioritization and selection models in \cite{farris2018vulcon, shah2019vulnerability, hore2022towards} directly maximize the cumulative utility or exposure scores of their respective factors to obtain the sets of prioritized vulnerability instances. Such a set of vulnerabilities may not contain all the important vulnerabilities, as their formulations do not maximize the average value of the selected vulnerabilities. In our proposed formulation, we counter this issue by maximizing the average of the cumulative value of all the attributes across all selected vulnerability instances subject to the total time available for mitigation in any given time-period. In addition, we take into consideration the largest set of attributes associated with any vulnerability and its respective host machine in published literature. Below, we present the input parameters, decision variables, objective function, constraints, and the output of the vulnerability selection model.

\noindent \textbf{Input parameters:}
\begin{itemize}
\item The attribute scores for all vulnerability instances, $v_{ij} ~\forall i,j $.
\item Expected time taken to mitigate a vulnerability instance $j$, $S_{j}$.
\item Total number of vulnerability instances in the scan report, $J$.
\item Total resources available at time $t$ (action from the DRL agent), $a_t$.
\end{itemize}

\noindent \textbf{Decision variables:}
\begin{itemize}
\item $z_{j}=1$ if vulnerability instance $j$ is selected, and $0$ otherwise.
\end{itemize}

\noindent \textbf{Objective function:}
The objective of the model is to select the set of vulnerability instances prioritized for mitigation that maximizes the average of the cumulative value of the attribute scores across all selected vulnerability instances. The objective function is given by:

\begin{equation}
y = Max~~ \frac{\sum_{j=1}^{J}\sum_{i=1}^{I}  v_{ij} * z_{j}}{\sum_{j=1}^{J}z_j} 
\label{eq5}
\end{equation}

\noindent \textbf{Constraint:}
The constraint for the model is the availability of resource time, $a_t$, at any given time $t$, which is obtained from the DRL agent. The constraint for the total time taken to mitigate the selected vulnerability instances not being higher than the total resource time available at time $t$ is expressed as:
\begin{equation}
\sum_{j=1}^{J} S_{j} * z_{j} \leq a_t 
\label{eq6}
\end{equation}

\noindent \textbf{Output:}
The output of the vulnerability prioritization and selection model is the set of prioritized vulnerability instances selected for mitigation.

\section{Numerical Experiments}
We worked closely with a CSOC to collect the vulnerability data and other relevant computer network information. Our conversations with the security analysts helped us determine various parameter values that were used in setting up the environment, and training and testing the proposed Deep VULMAN framework.

\subsection{Data Collection and Simulation Environment for CSOC Operations}
We developed a simulator from the real-world data set that we collected by working with a CSOC. We used two applications: Tenable's Nessus vulnerability scanner and Lansweeper to collect the vulnerability data. We collected a total of 98,842 vulnerability instances over a span of two years. We also collected relevant host machine data and alert data generated by the IDS. The Lansweeper report contained information about the host machines in the network, which included the software versions of the operating system and SQL server, among others. In this research study, we also integrated information from the IDS alert logs to obtain the intrusion status of the host machine. If a host machine with the reported vulnerability was identified in the IDS alert log for the respective time-period (say, between time $t-1$ and $t$), then this information was recorded and the intrusion status attribute of a vulnerability instance was set accordingly. All the machine-specific information for the host machines on which the vulnerabilities were found was added to the consolidated data set. The aggregated data set contained information about the host machine and vulnerability instances such as the host IP, CVE code, the description, the CVSS severity score, the importance of the machine in the network, the versions of software running on the host machine, and the estimated personnel-hours required to mitigate the vulnerability instance~\cite{farris2018vulcon}, among other known information. We then applied vulnerability data preprocessing techniques, which included quantification of the attributes: asset criticality, level of protection, and organizational relevance of the host machine, along with the CVSS severity of the vulnerability. We used the same categories and the quantification process as used in~\cite{farris2018vulcon} and \cite{hore2022towards}. 

We created an agent-based DES that mimics the arrival and mitigation process of vulnerability instances in a CSOC. With the help of a simulation model, we generated diverse patterns of new vulnerability arrivals to expose the DRL agent to uncertainty it may find in a real-world environment. From our discussions with the CSOC security personnel and historical evidence, along with the information published in literature~\cite{farris2018vulcon}, we modeled the vulnerability instance arrival process using a Poisson distribution and varied the average number of arrivals from 40 to 600 per week (indicating a very large network). We segregated our arrivals into three different categories, namely, high, medium, and low. Different patterns of arrivals, based on the aforementioned average numbers per week, were simulated for training the DRL agent. Some examples of arrival patterns for four consecutive weeks in a month include [high, high, low, low], [medium, medium, medium, low], and [low, high, medium, high], among others. Vulnerability instances were randomly sampled from the data set based on the arrival pattern (Poisson distribution with the respective average number of arrivals) at each time-step (i.e., $t = 1$ week) emulating the arrival process in the CSOC. All the information about the vulnerability instances is then passed to the decision-support component (explained in the next sub-section) to obtain an action indicating the set of vulnerability instances that are selected for mitigation. Upon receiving this information, the simulation algorithm is stepped forward and the selected vulnerability instances are mitigated utilizing the time assigned to each of the vulnerability instances in the consolidated data set. The total mitigation time of the selected vulnerability instances is then deducted from the available resource time from the previous time-step and the remaining resource time is carried forward to the next time-step. Each week is represented as a time-step in the simulator. Next, we describe the training and testing phases of the proposed Deep VULMAN framework.

\subsection{Training Phase}
We conducted our experiments with some of the hyper-parameter values from published literature \cite{schulman2017proximal} and \cite{9018554},  and tuned the remaining by trial-and-error, which involved running experiments with different sets of values. The PPO approach is known to be more forgiving to sub-optimal initialization of hyper-parameter values. We conducted the experiments on a machine with $11^{th}$ Gen Intel Core i7-12700H processor with NVIDIA GeForce RTX 2080 graphics card (16GB RAM).

We used a multi-layer perceptron (MLP) model with two hidden layers, each containing 68 perceptrons and Tanh activation functions for the actor and critic networks. It is to be noted that we implemented various architectures with a larger number of hidden layers and perceptrons but did not find any significant improvements in the performance of the DRL agent and selected the two hidden layer model that was computationally efficient among the others. The DRL agent took actions using the policy network. There was some standard deviation added to these actions, which started with a value of 0.65 and decayed to 0.01 with a rate of 0.025.  The decay rate and decay frequency are problem-specific, and hence we had to tune it with a trial-and-error approach. To avoid getting stuck in a local optimum and encourage exploration, we used 0.01 as the entropy co-efficient value, which was multiplied by the entropy and subtracted from the loss function. The value of the entropy factor was adopted from the literature~\cite{schulman2017proximal}. We set the maximum number of time-steps for training to 200M. At each time-step, the output of the DRL agent is provided as an input to the vulnerability prioritization and selection mathematical model and the vulnerability instances are selected for mitigation, which are then passed on to the CSOC operations environment. Based on these actions, a scalar reward value is calculated, which is derived from the two terms in the reward function (as shown in Equation~\ref{eq2}). We considered equal values of the weights used for the two reward terms in the reward function (in Equation~\ref{eq2}) and assigned a value of $10^{-5}$ to the cost per unit resource utilized ($C$). 

\subsection{Testing Phase}
We evaluated the DRL-enabled Deep VULMAN framework with the real-world vulnerability data from the collaborating CSOC. We set the standard deviation to zero during the testing phase, to avoid any further exploration by the DRL agent when taking actions using the policy network. We compared our method with two recent vulnerability selection methods from published literature, namely, VPSS~\cite{hore2022towards} and VULCON~\cite{farris2018vulcon}. We did not consider the CVSS-value based selection method in our comparison due to its limitation in taking the context of an organization into consideration. To compare the three approaches, we recorded the vulnerabilities that were selected for mitigation from (a) high value assets, (b) machines with lower level of protection, (c) organizationally relevant machines (i.e., web and database servers), and (d) machines with intrusion detection alert signals. Next, we describe and analyze the results obtained from the aforementioned experiments.

\section{Analysis of Results}

\begin{figure}[H]
\centering
\includegraphics[width=0.6\textwidth]{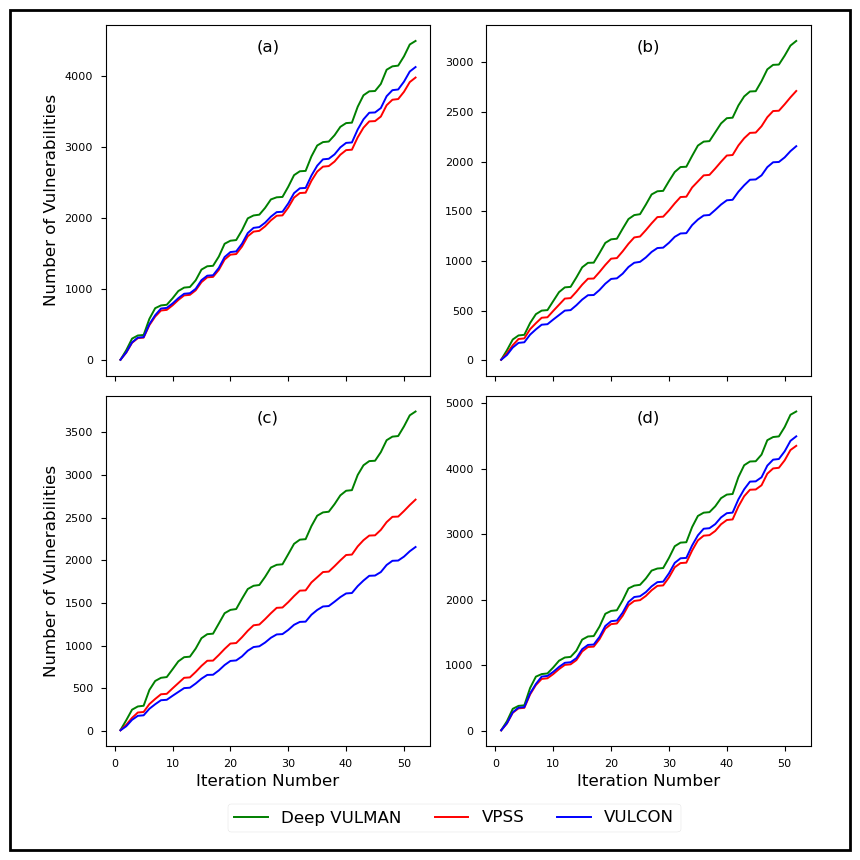}
\caption{Comparison of the total number of vulnerabilities selected from real-world data (one year) from (a) high value assets, (b)
machines with low level of protection, (c) organization-specific relevant machines, and (d) machines with intrusion alert signals.} 
\label{real_scenario}
\end{figure}

In this section, we present the evaluation results obtained using the real-world CSOC data. We evaluated the performance of our approach on the real-world vulnerability data set, which was collected from a collaborating CSOC for a period of one year. As shown in, Figure~\ref{real_scenario}(a) and Figure~\ref{real_scenario}(c), with our proposed approach more vulnerabilities are prioritized for mitigation from the important machines, i.e., web and database servers. We had observed similar results on the previously unseen simulated data. These results matched the requirements we had gathered from the security personnel at the CSOC. Figure~\ref{real_scenario}(d) shows another interesting result obtained using our method is the prioritization of vulnerabilities that were found on machines identified in potential attacks using the IDS alert data. Further investigation of these machines also revealed that the majority of these machines were identified to have a lower level of protection (old software versions with no or limited support from the vendor) Figure~\ref{real_scenario}(b), which indicates that they were an easier target for adversaries and vulnerabilities found in them must be prioritized. The results point towards a high degree of robustness an organization can achieve by employing the proposed DRL-enabled Deep VULMAN framework in the vulnerability triage process.

\begin{figure}[H]
\centering
\includegraphics[width=0.6\textwidth]{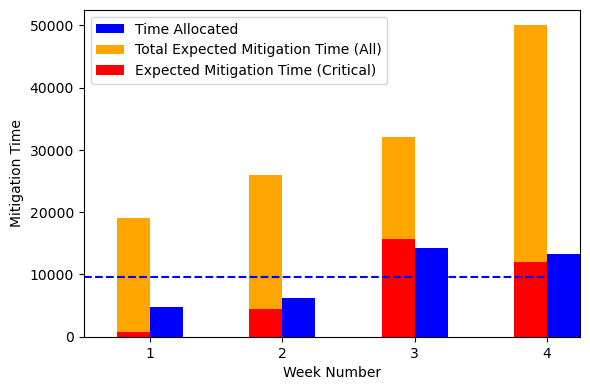}
\caption{Comparison between expected mitigation time of critical vulnerabilities and mitigation time allocated by the DRL agent.} 
\label{week_example}
\end{figure}

Figure~\ref{week_example} shows a particular episode (month), in which the vulnerability arrival pattern fluctuates between high, medium and low among the four time-steps (weeks). The orange bar shows the total expected mitigation time required (in minutes) to mitigate all the vulnerabilities identified in the network and the blue bar shows the amount of resources allocated by the DRL agent. We have highlighted the expected mitigation time of vulnerabilities, whose average for the cumulative normalized attribute values is high, in red. In particular, we have considered the vulnerability instances with the value of $\frac{\sum_{i=1}^{I} v_{i,j}}{I} \geq 0.75$ to show the effectiveness of our proposed approach in allocating resources to mitigate these \emph{critical} vulnerabilities. The dotted line in the figure represents the even distribution of resources, which is a commonly employed practice at the CSOCs and is utilized by the other two methods (VPSS and VULCON). It can be seen that the DRL agent allocates a lower than average number of resources in the first week to match the arrival pattern of vulnerabilities, followed by a lower than average number of resources in the second week, thereby saving more resources in anticipation of a larger number of new vulnerability arrivals in the last two weeks of the month. In this case, the DRL agent demonstrates that it is able to react appropriately in the first two time-steps by allocating less number of resources and has learned non-trivial decisions of utilizing conserved resources to counter an anticipated future event (of high arrivals). Accordingly, the prioritization and selection model is able to prioritize the selection of vulnerabilities across all the factors. 
This episode example amongst many others shows that the DRL agent has learned to make better decisions in the wake of uncertain vulnerability arrivals.

\section{Conclusions}
The paper presented a novel cyber vulnerability management framework, Deep VULMAN, to identify and prioritize important vulnerabilities for mitigation in the wake of uncertain vulnerability arrivals in a resource-constrained environment. We first trained a state-of-the-art DRL agent using a simulated CSOC operations environment, which was built using real-world CSOC data, to learn the near-optimal policy of allocating resources for selecting vulnerabilities for mitigation. Next, a mathematical model for vulnerability prioritization and selection was formulated and solved using the integer programming method, which generated the set of important vulnerabilities prioritized for mitigation based on the resources allocated. We conducted our experiments on both simulated and real-world vulnerability data for a one-year period. The results showed that our proposed framework outperformed the current methods by prioritizing the selection of maximum number of vulnerability instances from high-value assets, organizationally relevant machines (web and database servers), machines identified in intrusion detection alert signals, and machines with lower level of protection. The DRL agent learned non-trivial decisions in the wake of uncertain vulnerability arrival patterns. For instance, the agent was able to anticipate future events of high vulnerability arrivals, and accordingly adjusted (conserved) the allocation of resources in earlier time-steps to counter the important vulnerabilities during those events.

We first trained a state-of-the-art DRL agent using a simulated CSOC operations environment, which was built using real-world CSOC data, to learn the near-optimal policy of allocating resources for selecting vulnerabilities for mitigation. Next, a mathematical model for vulnerability prioritization and selection was formulated and solved using the integer programming method to obtain the prioritized set of important vulnerabilities selected for mitigation. We conducted our experiments on both simulated and real-world vulnerability data for a one-year period. The results showed that our proposed framework outperformed the current methods by prioritizing the selection of the maximum number of vulnerability instances from high-value assets, organizationally relevant machines (web and database servers), machines identified in intrusion detection alert signals, and machines with lower level of protection. The DRL agent learned non-trivial decisions in the wake of uncertain vulnerability arrival patterns. For instance, the agent was able to anticipate future events of high vulnerability arrivals, and accordingly adjusted (conserved) the allocation of resources in earlier time-steps to counter the important vulnerabilities during those events.

The proposed DRL-enabled cyber vulnerability management framework, Deep VULMAN, can strengthen the security posture of an organization by generating robust policies in uncertain and resource-constrained real-world environments. In this study, we also determined the optimal allocation of limited number of resources that are available in a CSOC, across different time-steps under uncertainty. An interesting follow-up work or a future research direction can include the development of data-driven models to determine an optimal number of security personnel needed to achieve the performance goal of a vulnerability management team. Furthermore, a trade-off study can be conducted comparing the impact of budget on  staffing and performance of the vulnerability management teams.

\section*{Acknowledgments}
This work was supported in part by the U.S. Military Academy (USMA) under Cooperative Agreement No. W911NF-22-2-0045, the U.S. Army Combat Capabilities Development Command (DEVCOM) Army Research Laboratory under Support Agreement No. USMA21050, and the U.S. Army DEVCOM C5ISR Center under Support Agreement No. USMA21056. The views and conclusions expressed in this paper are those of the authors and do not reflect the official policy or position of the U.S. Military Academy, U.S. Army, U.S. Department of Defense, or U.S. Government.

\bibliographystyle{unsrtnat}
\bibliography{references}  






\end{document}